\documentclass[
	a4paper, 
	10pt, 
	unnumberedsections, 
	twoside, 
]{LTJournalArticle}
\usepackage[
backend=biber, 
citestyle=numeric, 
bibstyle=numeric, 
sorting=anyt, 
]{biblatex}

\runninghead{VN-HGCN} 

\footertext{} 

\title{Virtual Nodes based Heterogeneous Graph Convolutional Neural Network for Efficient Long-Range Information Aggregation} 

\author{%
	Ranhui Yan\textsuperscript{1}, Jia Cai\textsuperscript{2}\thanks{Corresponding author: \href{mailto:jiacai1999@gdufe.edu.cn}{jiacai1999@gdufe.edu.cn} \\ The work described in this paper was supported partially by the National Natural Science Foundation of China (12271111), Guangdong Basic and Applied Basic Research Foundation (2022A1515011726), Special Support Plan for High-Level Talents of Guangdong Province (2019TQ05X571). \\ Received by The 33rd International Conference on Artificial Neural Networks.}
}

\date{\footnotesize\textsuperscript{\textbf{1}}School of Statistics and Mathematics, Guangdong University of Finance \& Economics, Guangzhou, China \\ \textsuperscript{\textbf{2}}School of Digital Economics, Guangdong University of Finance \& Economics, Guangzhou, China}

\renewcommand{\maketitlehookd}{%
	\begin{abstract}
		\noindent Heterogeneous Graph Neural Networks (HGNNs) have exhibited powerful performance in heterogeneous graph learning by aggregating information from various types of nodes and edges. However, existing heterogeneous graph models often struggle to capture long-range information or necessitate stacking numerous layers to learn such dependencies, resulting in high computational complexity and encountering over-smoothing issues.  In this paper, we propose a {\bf V}irtual {\bf N}odes based {\bf H}eterogeneous {\bf G}raph {\bf C}onvolutional {\bf N}etwork (VN-HGCN), which leverages virtual nodes to facilitate enhanced information flow within the graph. Virtual nodes are auxiliary nodes interconnected with all nodes of a specific type in the graph, facilitating efficient aggregation of long-range information across different types of nodes and edges. By incorporating virtual nodes into the graph structure, VN-HGCN achieves effective information aggregation with only $4$ layers. Additionally, we demonstrate that VN-HGCN can serve as a versatile framework that can be seamlessly applied to other HGNN models, showcasing its generalizability. Empirical evaluations validate the effectiveness of VN-HGCN, and extensive experiments conducted on three real-world heterogeneous graph datasets demonstrate  the superiority of our model over several state-of-the-art baselines.
	\end{abstract}
}

\begin{document}

\maketitle 


\section{Introduction}
Graph-structured data, such as citation networks and social networks, are ubiquitous in the real world.
Graph neural networks (GNNs) have emerged as a powerful deep learning technique capable of capturing intricate relationships and dependencies between nodes and edges.  GNNs have been widely used in various fields, such as natural language processing (NLP), computer vision (CV), recommender systems, and bioinformatics.
However, most existing GNNs mainly focus on dealing with homogeneous graphs, where all the nodes and edges belong to the same type.
\begin{figure}[htbp]
	\centering
	\includegraphics[width=0.9\linewidth]{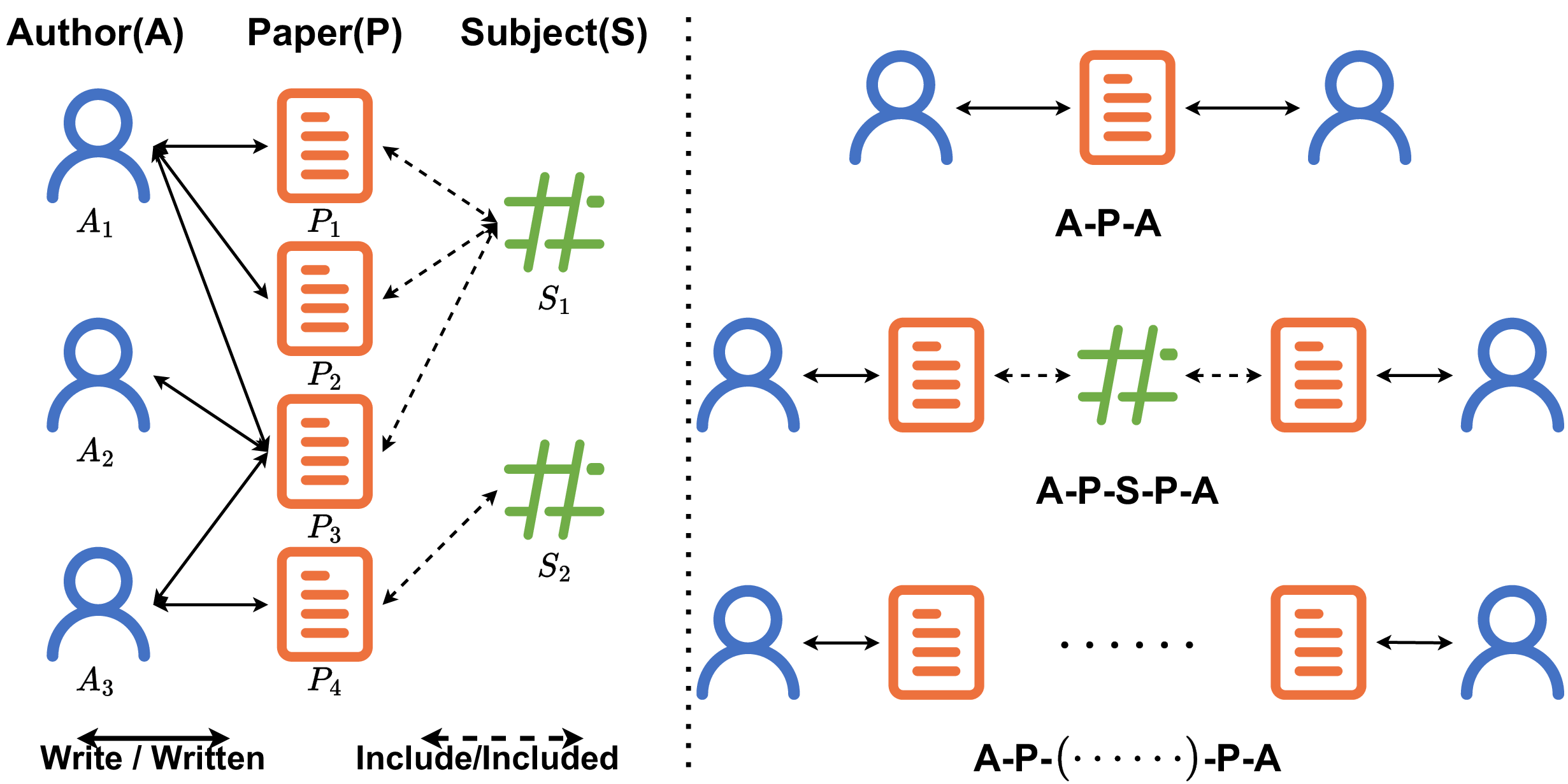}
	\caption{An example of heterogeneous graph (left) and the meta-path (right).}
	\label{example}
\end{figure}
In practice, many datasets exhibit heterogeneous graph structures containing diverse types of nodes or edges.
For instance, as shown in Fig. \ref{example}, an academic citation network (ACM) contains three distinct types of nodes: author, paper, and subject; and two pairs of edge connections: write/written and include/included between author-paper (A-P) and subject-paper (S-P) respectively.
Meta-paths \cite{metapath} are  predefined sequences of nodes and edges utilized to exploit different semantics and describe the semantic relations between nodes in a heterogeneous graph. As shown in Fig. \ref{example}, the ACM dataset usually defines meta-paths \textit{APA} and \textit{APSPA} before applying graph representation models.
The meta-path \textit{APA} reveals the semantics of co-authorship of papers, while \textit{APSPA} denotes that two authors conduct research in the same field.

Previous works leverage meta-paths for message passing to aggregate semantic information from heterogeneous graphs. These methods, including metapath2vec \cite{mp2vec}, HAN \cite{HAN}, MAGNN \cite{MAGNN}, and HPN \cite{HPN}, typically extract rich semantics and expressive node embeddings. However, they require expert prior knowledge to define meta-paths, and their effectiveness is comparable to homogeneous graph neural networks such as GCN \cite{gcn} and GAT \cite{gat} (heterogeneous graph extension). Despite this, missing key meta-paths can lead to significant performance degradation, as observed in the HAN model when \textit{APA} is absent.  GTN \cite{CPGCNLUO} addressed the importance of meta-paths within a specific length by combining adjacent matrices and performing GCN. However, GTN's complexity significantly escalates with more layers or specified lengths, resulting in high computational complexity due to the exponential growth in the number of meta-paths for different node/edge types.

To address this challenge, several studies have designed heterogeneous graph neural networks without predefined meta-paths.
JUST \cite{JUST} employed random walks via jump and stay strategies to mitigate bias towards highly visible domains in heterogeneous graphs. Nontheless,  a more common paradigm is to project nodes of different types into a shared semantic space for message passing or convolutional operations, as demonstrated in HetSANN \cite{HetSANN}, HGT \cite{HGT}, and SimpleHGN \cite{SimpleHGN}. Additionally, experimental results from ie-HGCN \cite{iehgcn} indicate that the most important meta-path may not always align with human prior knowledge.

Furthermore, to learn more informative node embeddings, long-range information must be considered. While stacking more layers theoretically increases the receptive field of the model and captures long-range information, it also leads to severe over-smoothing problems \cite{HPN,oversm} in homogeneous graphs. As the number of layers increases, node embeddings become indistinguishable,  resulting in performance degradation. Therefore, vanilla HGNNs struggle to integrate long-range information, aggregating neighbors recursively in each layer. JK-Net \cite{jknet} and HPN \cite{HPN} employed residual/dense connections like DenseNet \cite{densenet} and ResNet \cite{resnet} to alleviate over-smoothing issue. However, they require stacking $k$ layers to aggregate neighbors with $k$ hops away, significantly increasing computational complexity, especially for large $k$ values.  Overall, existing heterogeneous graph models may confront with the following three challenges: (1). How to deal with over-smoothing without expert prior knowledge and meta-path. (2). How to capture long-range information. (3). How to avoid stacking too many layers, which may lead to dramatic increasing of the time complexity.

Inspired by the strategy of adding virtual nodes to facilitate message passing \cite{vn}, we incorporate virtual nodes into heterogeneous graphs, enabling message passing over long distances and reducing the need for stacked layers. We propose {\bf V}irtual {\bf N}ode-based {\bf H}eterogeneous {\bf G}raph {\bf C}onvolutional {\bf N}etwork (VN-HGCN) to address the aforementioned challenges simultaneously, capable of learning long-range dependencies with only $4$ layers without predefined meta-paths. Moreover, the virtual nodes adding strategy can be effortlessly implemented as a universal framework for existing models, recursively aggregating neighbors to update node embeddings in models like HGT,  HetSANN, SimpleHGN, and ie-HGCN. 
We also provide empirical verification of the long-range information integration on the node embeddings.
The proposed VN-HGCN  does not require predefined meta-paths, but rather aggregates information along any possible path within a given length as done in HGT, HetSANN, SimpleHGN, and ie-HGCN.
Experiments on node classification task demonstrate the effectiveness of VN-HGCN.
Overall, the main contributions of this paper are addressed as follows:
\begin{itemize}
	\item We propose VN-HGCN, an efficient and long-range information-aware heterogeneous graph model.
	\item VN-HGCN serves as a general universal framework applicable to other heterogeneous graph models, and its effectiveness is validated.
	\item Experimental results illustrate the superiority of VN-HGCN over various state-of-the-art models, with parameter sensitivity analysis demonstrating the robustness of the proposed model.
\end{itemize}

\section{Preliminaries and Related Work}
In this section, we introduce some basic definitions and long-range information extraction related work.
\subsection{Heterogeneous graph}
Heterogeneous graph, known as heterogeneous Information Network (HIN), consists of more than two types of node (object) or edge (relation), which makes it a more realistic representation of graph structural data in the real world.
\textbf{Heterogeneous Graph \cite{hin}}. A heterogeneous graph is denoted as $\mathcal{G} = (\mathcal{V,E},\phi,\psi)$, where $\mathcal{V}$ and $\mathcal{E}$ are the sets of nodes and edges, respectively, $\phi:\mathcal{V}\to \mathcal{A}$ and $\psi:\mathcal{E}\to \mathcal{R}$ are mapping functions that assign a node type and an edge type (relation) to each node and edge, respectively, $\mathcal{A}$ and $\mathcal{R}$ are the sets of node and edge types, respectively,  where $|\mathcal{A}|+|\mathcal{R}|>2$.
Furthermore, each heterogeneous graph can be abstract as a meta template called \textbf{Network Schema}.

\textbf{Network Schema \cite{hin}}. The network schema is meta template for a heterogeneous graph $G = (\mathcal{V,E},\phi,\psi)$ , which is denoted as $T_G(\mathcal{A}, \mathcal{R})$. It is a graph defined over node types $\mathcal{A}$ with edges from $\mathcal{R}$.

\subsection{Long-Range Information Extraction}
While HGNNs have shown promising results, they often struggle to capture long-range information effectively.  Models like GCN and RGCN theoretically have the capability to aggregate information from $k$-th order neighbors by stacking multiple layers. However, as the number of layers increases, they encounter issues such as vanishing gradients and over-smoothing (referred to as semantic confusion in heterogeneous graphs), resulting in nodes becoming indistinguishable and leading to performance degradation.

To overcome these challenges,  various approaches have been explored. Some studies have adopted residual or dense connections, inspired by techniques from computer vision \cite{resnet,densenet}, in GNN architectures like DeepGCNs \cite{deepgcns}, DeeperGNN \cite{deepergnn}, JK-Net \cite{jknet}, and EIGNN \cite{eignn}.  Others, such as Luan et al. \cite{sb} integrated multi-scale information to enhance node embeddings. Models like MGNNI \cite{mgnni} focused on aggregating multi-hop neighbors in each layer to extend the effective range of message passing.  CP-GCN \cite{CPGCNLUO} adopted the context path to capture the high-order relationship information and introduced the context path probability to model the learning objective function. For heterogeneous graph,  HPN \cite{HPN} employed a residual-like architecture to mitigate over-smoothing in heterogeneous graphs. Despite their expressive power and ability to aggregate long-range nodes, these models often require stacking numerous layers, leading to high computational complexity.

Inspired by the use of virtual nodes in graph pooling and graph prediction \cite{vn,GCVN,GWM}, we assign hierarchical virtual nodes to facilitate long-range information flow across the graph. With the assistance of virtual nodes, the proposed VN-HGCN can effectively explore $k$-hop neighbors of a target node within $4$ layers.

\section{Virtual Nodes Based Heterogeneous Graph Convolutional Network (VN-HGCN)}
In this section, we elucidate the basic idea of the proposed framework. 
VN-HGCN, with its $4$ layers, facilitates message passing between nodes situated far apart in the graph. The key steps involved in computing node embeddings with VN-HGCN include adding virtual nodes, transformation, and aggregation.  Let's delve into each of these steps in detail.
\begin{figure}[htbp]
	\centering
	\includegraphics[width=0.8\linewidth]{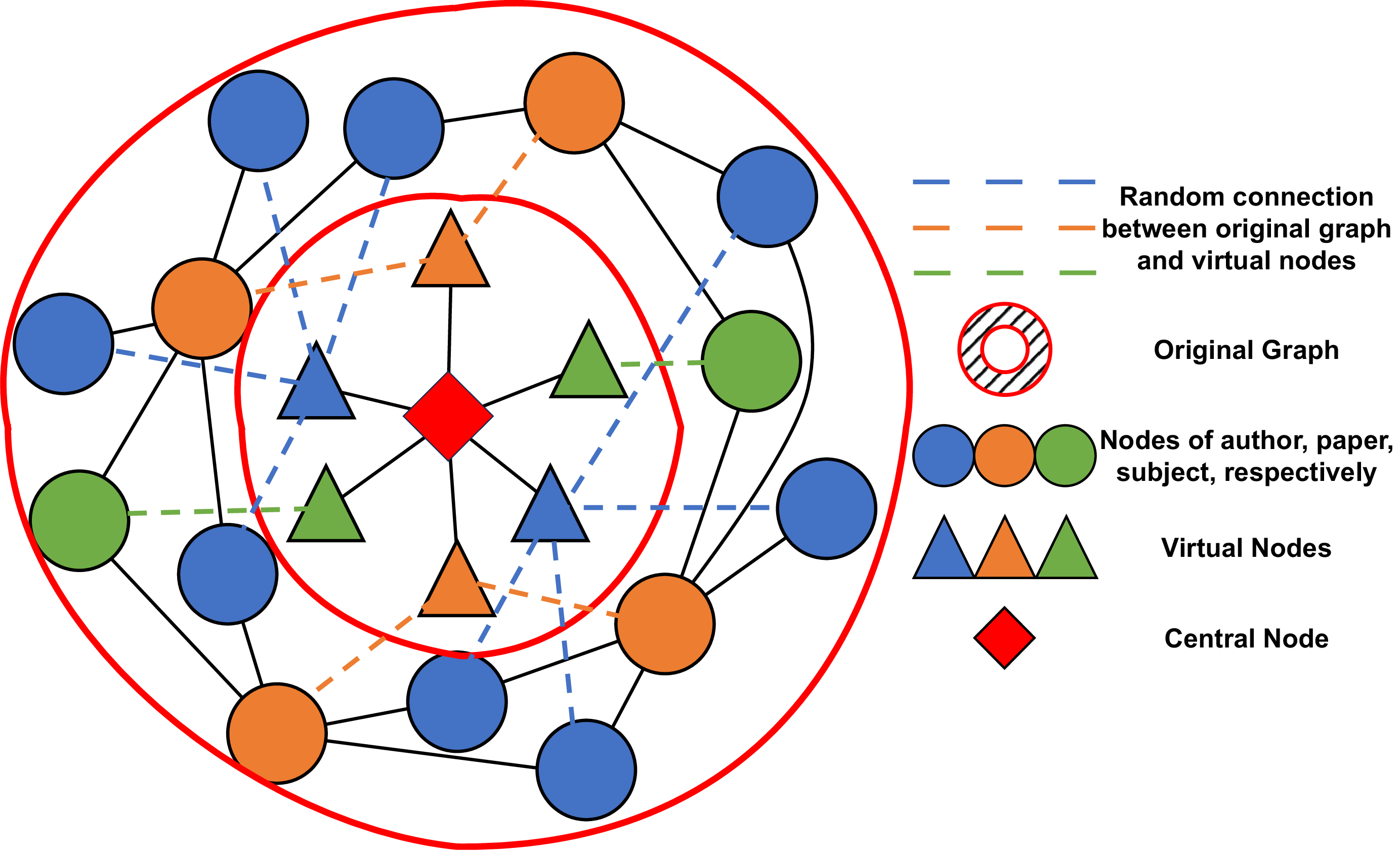}
	\caption{Adding 2 virtual nodes for each type in the ACM dataset.}
	\label{add_vn}
\end{figure}

\subsection{Adding Virtual Nodes}
Before performing message passing, we introduce hierarchical virtual nodes to the original heterogeneous graph.
First, we assign type-level virtual nodes for each node, randomly connecting each node of a certain type with one of these virtual nodes.
For example, in an ACM citation network each `author' node is randomly connected with one of the `author' virtual nodes, and each `paper' node with one of the `paper' virtual nodes, and so forth  (see Fig. \ref{add_vn}).
Through these connections, virtual nodes can absorb all the feature information from distinct types of nodes, effectively pooling the information of the graph into several type-level virtual nodes.
Subsequently, we enable these type-level virtual nodes to exchange information with each other, facilitating inter-type virtual node message passing.
We introduce a central virtual node to connect with type-level virtual nodes.
For instance, as shown in Fig. \ref{add_vn}, `author' virtual nodes  are connected with every `author' node, while the central node is connected with all the type-level virtual nodes.
\begin{figure}
	\centering
	\includegraphics[width=.5\linewidth]{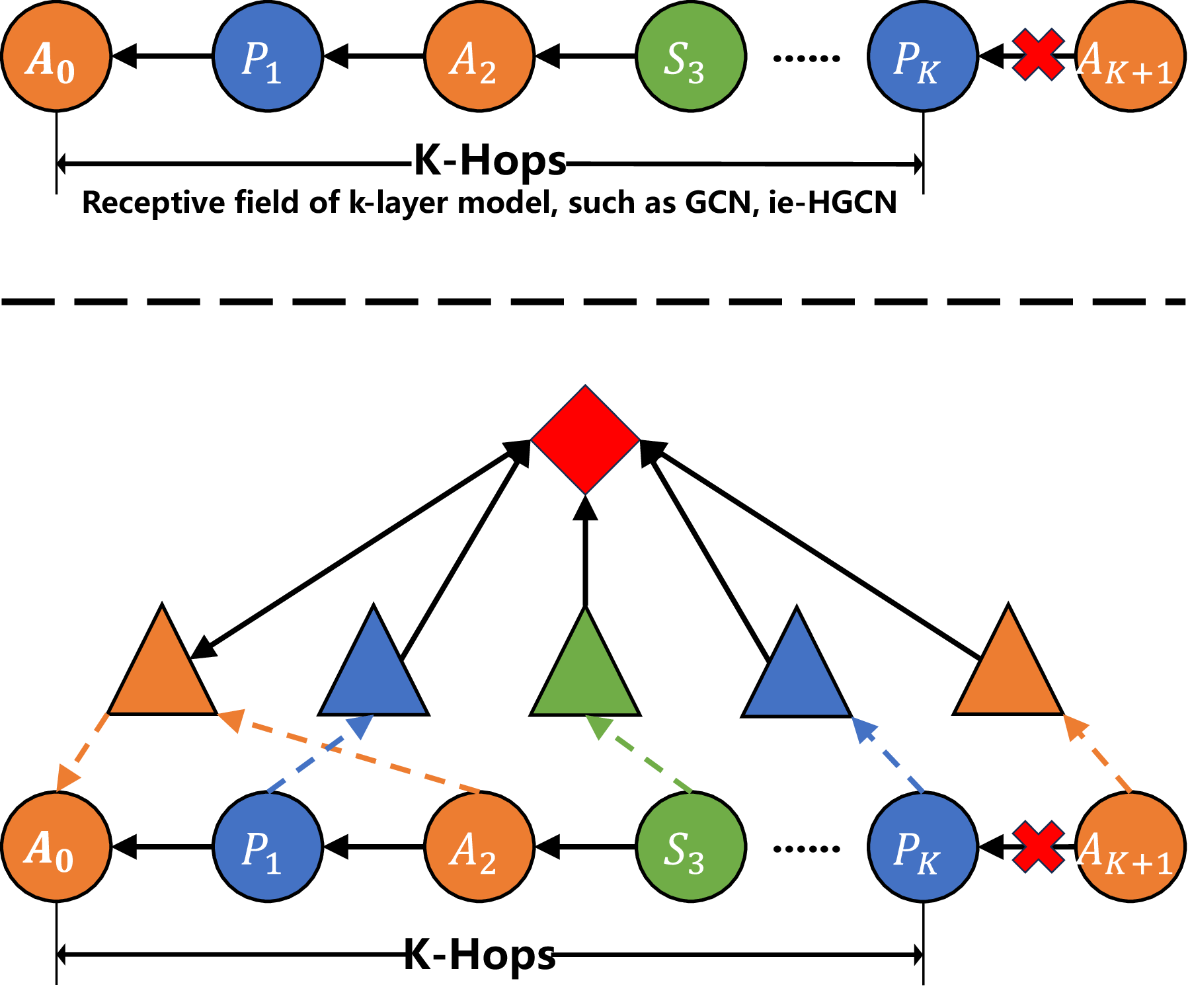}
	\caption{Message passing before (up) and after (down) adding virtual nodes.}
	\label{mp}
\end{figure}

After augmenting the original graph structure by adding virtual nodes, message passing is no longer confined to the original network schema \cite{metapath}.
Instead, it follows the new network schema that includes virtual nodes.
Moreover, the number of node types changes from $|\mathcal{A}|$ to $(2|\mathcal{A}|+1)$.
With the help of virtual nodes, every node pair $(v_i,v_j)$ in the graph can be connected by a path of length $4$, regardless of the distance: $v_i-V_{\phi(v_i)}-C-V_{\phi(v_j)}-v_j$.
As illustrated in Fig. \ref{mp}, for a node of `author' $A_0$ and its $K$-hops neighbor node of `author' $A_{K+1}$, the message passing between them depends on the `$A_0-P_1-\cdots-P_{K}-A_{K+1}$' path of $K$ lengths before adding virtual nodes.
After adding virtual nodes, the distance between $A_0$ and $A_{k+1}$ is shortened to $4$.
To initialize type-level virtual nodes, we utilize the mean of node features connected with the specific type-level virtual nodes.
The central virtual node is initialized with an all-ones vector.
Additionally, to mitigate the risk of overfitting caused by connecting too many nodes with type-level virtual nodes, we use drop-edge \cite{de}.

\subsection{Transformation}
To process the node aggregation,  transformation is required to project each node type into a common semantic feature space. Specifically, we use a set of learnable matrices $W^{(\ell)}$ to perform the transformation, yielding transformed node embeddings in the $\ell$-th layer:
\begin{equation}\label{eq1}
	Y^{(\ell)}_{\Lambda_i} = H^{(\ell)}_{\Lambda_i} W^{(\ell)}_{\Lambda_i},
\end{equation}
\begin{equation}\label{eq2}
	Y^{(\ell)}_{\Lambda_j\to\Lambda_i} = H^{(\ell)}_{\Lambda_j} W^{(\ell)}_{\Lambda_j\to\Lambda_i},
\end{equation}
where $H^{(\ell)}_{\Lambda_i}$ and $H^{(\ell)}_{\Lambda_j}$ are the node embedding matrices of type ${\Lambda_i}$ and ${\Lambda_j}$, respectively.
When $\ell=0$, $H^{(\ell)}_{\Lambda_i}=X_{\Lambda_i}$, where $X_{\Lambda_i}$ is the original feature matrix of type $\Lambda_i$.

\subsection{Aggregation}
After transformation, VN-HGCN updates node embeddings via two types of aggregation: node-level aggregation and type-level aggregation.

{\bf Node-Level Aggregation.}
Note that adjacent nodes of target node come from different types, inspired by ie-HGCN, we utilize normalized adjacency matrices $\hat{A}$ to perform node-level aggregation for efficiency and interpretability.
For target $\Lambda_i$, the normalized adjacency matrix of $\Lambda_i$ to $\Lambda_j$ is $\hat{A}_{(\Lambda_i,\Lambda_j)}=D^{-1}_{(\Lambda_i,\Lambda_j)}A_{(\Lambda_i,\Lambda_j)}$, where $D_{(\Lambda_i,\Lambda_j)}\in \mathcal{R}^{|\mathcal{V}_{\Lambda_i}|\times |\mathcal{V}_{\Lambda_j}|}$ is the degree matrix of $A_{(\Lambda_i,\Lambda_j)}$, $|\mathcal{V}_{\Lambda_i}|$ is the number of node type $\Lambda_i$.
Then, we perform node-level aggregation to obtain the embeddings of the target and other types of nodes:
\begin{equation}\label{eq3}
	Z^{(\ell)}_{\Lambda_i} =Y^{(\ell)}_{\Lambda_i},
\end{equation}
\begin{equation}\label{eq4}
	Z^{(\ell)}_{\Lambda_j\to\Lambda_i} = \hat{A}_{(\Lambda_i,\Lambda_j)} Y^{(\ell)}_{\Lambda_j\to\Lambda_i}, 
\end{equation}
where $Z^{(\ell)}_{\Lambda_j\to\Lambda_i}$ stores aggregated information of $\Lambda_j$ to $\Lambda_i$, i.e., the $i$-th row of $Z^{(\ell)}_{\Lambda_j\to{\Lambda_i}}$ stores the information of nodes that come from $\Lambda_j$ and connecting with the $i$-th node of $\Lambda_i$.

{\bf Type-Level Aggregation.}
To update the embeddings of  the target $H_{\Lambda_i}$ in the next layer, we aggregate the embeddings of $\Big(Z^{(\ell)}_{\Lambda_i},\{Z^{(\ell)}_{\Lambda_j\to{\Lambda_i}}\}\Big)$  via  an attention mechanism \cite{att}.
Specifically, we transform each embedding $Z$ and calculate the inner product with an attention vector $\mathbf{q_{\Lambda}}\in\mathcal{R}^{d_a \times 1}$, $\Lambda\in\mathcal{A}$ ($\mathbf{q}$ is initialized with an all-ones vector), followed by a nonlinear activation function such as ${\rm tanh}$:
\begin{equation}
	\mathbf{e}^{(\ell)}_{\Lambda_i}= {\rm tanh}[(Z^{(\ell)}_{\Lambda_i} E^{(\ell)}_{\Lambda_i})\mathbf{q}^{(\ell)}_{\Lambda_i}],
	\label{self_att}
\end{equation}
\begin{equation}
	\mathbf{e}^{(\ell)}_{\Lambda_j\to\Lambda_i}=  {\rm tanh} [(Z^{(\ell)}_{\Lambda_j\to\Lambda_i} E^{(\ell)}_{\Lambda_i})\mathbf{q}^{(\ell)}_{\Lambda_i}],
	\label{adj_att}
\end{equation}
where $E^{(\ell)}_{\Lambda_i}$ is the matrix of learnable parameters.
Subsequently, we normalize the resulting intra-type attention $\mathbf{e}^{(\ell)}_{\Lambda_i}$ and inter-type attention  $\mathbf{e}^{(\ell)}_{\Lambda_j\to\Lambda_i}$ using softmax function for numerical stability and explainability:
\begin{equation}
	\alpha^{(\ell)}_{\Lambda_i}=\frac{\exp(e^{(\ell)}_{\Lambda_i})}{\exp(e^{(\ell)}_{\Lambda_i})+\sum_{k}\exp(e^{(\ell)}_{\Lambda_k\to\Lambda_i})},
	\label{self_att_n}
\end{equation}
\begin{equation}
	\alpha^{(\ell)}_{\Lambda_j\to\Lambda_i}=\frac{\exp(e^{(\ell)}_{\Lambda_j\to\Lambda_i})}{\exp(e^{(\ell)}_{\Lambda_i})+\sum_{k}\exp(e^{(\ell)}_{\Lambda_k\to\Lambda_i})}.
	\label{adj_att_n}
\end{equation}
Finally,  we perform type-level aggregation with inter-type attention and update the node embeddings of the next layer:
\begin{equation}\label{eq9}
	H^{(\ell+1)}_{\Lambda_i} = \sigma\left[\alpha^{(\ell)}_{\Lambda_i} Z^{(\ell)}_{\Lambda_i}+\sum (\alpha^{(\ell)}_{\Lambda_j\to{\Lambda_i}}Z^{(\ell)}_{\Lambda_j\to{\Lambda_i}})\right],
\end{equation}
where $\sigma$ is a non-linear activation, we choose ReLU in this paper.

\begin{algorithm}
	\DontPrintSemicolon
	\SetAlgoLined
	\SetKwInOut{Input}{Input}\SetKwInOut{Output}{Output}
	\Input{Heterogeneous graph $\mathcal{G}$  = $(\mathcal{V,E}, \phi, \psi)$, Feature matrix $X$, Number of  type-level virtual nodes $N_V$, Number of layers $K$.}
	\Output{Node representation $Z$.}
	\BlankLine
	
	\tcp{Adding virtual nodes}
	\For{$\Lambda_i \in \mathcal{A}$}{
		Add $N_V$ type-level virtual nodes $V_{\Lambda_i}$ for $\Lambda_i$\;
		Build random connection between $V_{\Lambda_i}$ and $\Lambda_i$\;
		Set a central node $C$ and connect  with all of $V_{\Lambda_i}$\;
	}
	\tcp{Network schema changes, and $|\mathcal{A}| \leftarrow 2|\mathcal{A}|+1$}
	\For{$\ell=0,1,\cdots,k-1$}{
		\For{$\Lambda_i \in \mathcal{A}$}{
			Compute $Y^{(\ell)}_{\Lambda_i}$ via (\ref{eq1});\\
			Compute $Z^{(\ell)}_{\Lambda_i}$ via (\ref{eq3});\\
			\For{$\Lambda_j$ neighboring with $\Lambda_i$ in network schema}{
				Calculate $Y^{(\ell)}_{\Lambda_j\to\Lambda_i}$ by (\ref{eq2});\\
				Calculate $Z^{(\ell)}_{\Lambda_j\to\Lambda_i}$ by (\ref{eq4});\\
			}
			Calculate intra-type attention vector  $\mathbf{e}^{(\ell)}_{\Lambda_i}$ via (\ref{self_att});\\
			\For{$\Lambda_j$ neighboring with $\Lambda_i$ in network schema}{
				Calculate inter-type attention vector $\mathbf{e}^{(\ell)}_{\Lambda_j\to\Lambda_i}$ via (\ref{adj_att});\\
			}
			Compute  $\alpha^{(\ell)}_{\Lambda_i}$ and $\alpha^{(\ell)}_{\Lambda_j\to\Lambda_i}$ via (\ref{self_att_n}) and (\ref{adj_att_n}), respectively;\\
			Update node embeddings $H^{(\ell+1)}_{\Lambda_i}$ via (\ref{eq9}).
		}
		Update parameters according to loss function by gradient descent method;\\
		\Return Final node representation $H^K$;}
	\caption{VN-HGCN}
\end{algorithm}

\subsection{Model Analysis}
The model analysis of VN-HGCN is summarized as follows:
\begin{itemize}
	\item  Adding virtual nodes shortens the distance between nodes in the graph, enabling recursive aggregation of neighbors to capture long-range information. VN-HGCN, a heterogeneous graph convolutional network, is designed to learn node representations.
	\item The number of type-level virtual nodes is a key hyper-parameter for adjusting the graph scale and avoiding overfitting.
	\item  For a $N$ layer VN-HGCN, the total number of parameters is approximate to $[N( |\mathcal{A}| |\mathcal{V}|d'_{\Lambda_i} + (|\mathcal{A}|+|\mathcal{V}|)d_a)]$ (which is influenced by the specific dataset), where $d'_{\Lambda_i}$ and $d_a$ are the dimension of the shared feature space and the attention vector respectively. The number of parameters is linear to the number of nodes and node types, without significantly increasing compared to other HGNNs with similar computational processes.
	\item The time complexity of VN-HGCN primarily arises from the transformation and aggregation steps. In practice,  $d'_{\Lambda_i}$ and  $d_a$ are very small compared to the number of nodes and edges of the graph. Therefore, the time complexity of transformation is $\mathcal{O}\left(|\mathcal{V}|\right)$.
	The time complexity of aggregation is $\mathcal{O}\Big( |\mathcal{A}|(|\mathcal{E}|+ |\mathcal{V}|) \Big)$, which consists of attention computation process, the node-level and type-level aggregation.
	Overall, for VN-HGCN with $N$ layer, the time complexity is $\mathcal{O}\left(N|\mathcal{A}|(|\mathcal{V}|+|\mathcal{E}|)\right)$.
\end{itemize}
\begin{figure}
	\centering
	\includegraphics[width=\linewidth]{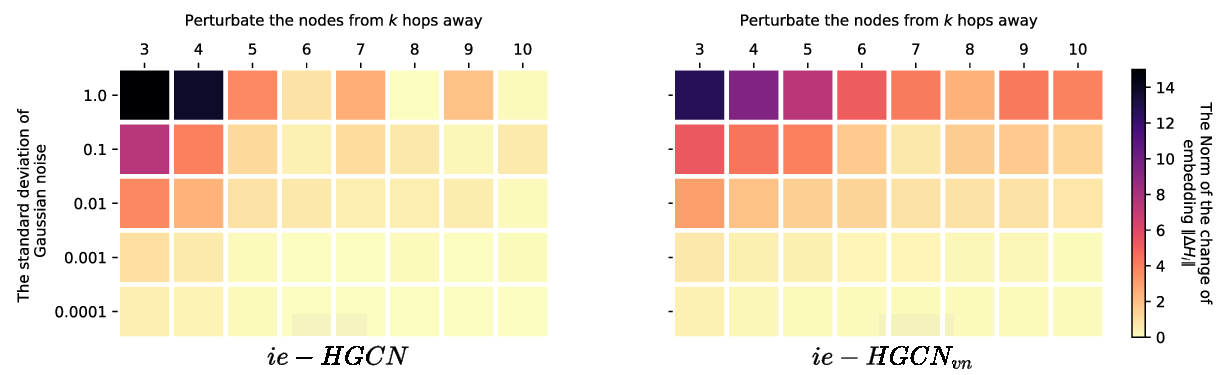}
	\caption{The norm of the change of embedding $\left\|\Delta H_i\right\|$ after perturbation.}
	\label{norm}
\end{figure}
\subsection{Scalability of the Model}
Intuitively, VN-HGCN presents a straightforward and adaptable solution that can be easily extended as a general framework for application in other heterogeneous graph neural networks, such as HGT, ie-HGCN, and others. To validate the effectiveness of virtual nodes, we conduct an experiment inspired by the exploration of node influence range in multi-scale graph neural network interpolation (MGNNI) \cite{mgnni}.  Specifically, we introduce virtual nodes into ie-HGCN (referred to as ie-HGCN$_{vn}$), a variant of ie-HGCN, and perturb nodes located at various distances ($3$ to $10$ hops) from a specific node using Gaussian noise with different variances.
Both ie-HGCN and its virtual node-based variant are limited to $4$ layers.

The experiment is conducted on the DBLP dataset, with the target node and the perturbed nodes belonging to the same node type to be estimated, i.e., `author'. The norm of the change in the learned node embedding $\left\|\Delta H_i\right\|$ is used to evaluate the influence of the k-hop neighbors.
As depicted in Fig. \ref{norm}, the darker color indicates a higher values of $\left\|\Delta H_i\right\|$. It's evident that the norm of the change decreases as the variance of the noise decreases. For ie-HGCN, the norm of the change decreases significantly as the distance between the node increases. However, for ie-HGCN$_{vn}$, the learned node embedding remains influenced by perturbed nodes even when they are $k$-hops away  (where $k$ is larger than the number of layers).
This observation suggests that ie-HGCN$_{vn}$ is more sensitive to perturbations from long-range distances than ie-HGCN, with the norm of the change consistently larger than that of ie-HGCN.
This implies that adding virtual nodes aids the model in exploring long-range neighbors and facilitating message passing.

\begin{table}
	\centering
	\caption{Statistics of the datasets.}
	\label{dataset_desc}
	\resizebox{\linewidth}{!}{
		\begin{tabular}{c|c|c|c|c|c|c}
			\hline
			Dataset   & Node Type    & Nodes & Edges                & Features & Classes & Meta-paths / Semantics                   \\ \hline
			& Author (A)   & 7167  & Paper-Author (13407) & 128      &         & P-A-P: papers by the same author         \\
			ACM       & Paper (P)    & 4017  & Paper-Subject(4019)  & 128      & 3       & P-S-P: papers in the same research field \\
			& Subject (S)  & 60    &                      & 128      &         &                                          \\ \hline
			& Author (A)   & 4057  & Paper-Author(19645)  & 334      &       & A-P-A: coauthor of a paper               \\
			DBLP & Paper (P)      & 14328 & Paper-Conference(14328) & 20   &4  & A-P-C-P-A: authors who submitted papers at the same conference \\
			& Conference (C) & 20    & Paper-Term(85810)       & 4231 &  & A-P-T-P-A: authors in the same research field                  \\
			& Term (T)     & 7723  &                      & 50       &         & \textbf{}                                \\ \hline
			& Actor (A)    & 5257  & Movie-Actor(12828)   & 3066     &         & M-D-M: movies by the same director       \\
			IMDB      & Director (D) & 2081  & Movie-Director(4278) & 3066     & 3        & M-A-M: movies with the same actor        \\
			\textbf{} & Movie (M)    & 4278  &                      & 3066     &       &                                          \\ \hline
	\end{tabular}}
\end{table}

\section{Experiment}
In this section, we conduct experiments on three real-world dataset: ACM, DBLP, and IMDB for node classification task\footnote{Code is available at \href{https://github.com/Yanrh1999/VN-HGCN}{https://github.com/Yanrh1999/VN-HGCN}}. 
\subsection{Setting}
The statistics of the datasets and the meta-paths used  are summarized in Table \ref{dataset_desc}.
Each dataset  is randomly divided into training,  validation,  and test sets according to varying ratios of the training data ($\{20\%$, $40\%$, $60\%$,  $80\%\}$) with the remaining data equally divided between the validation and test sets. 

We compare VN-HGCN with several state-of-the-art models implemented by OpenHGNN \cite{openhgnn}, including HAN, MAGNN,  HGT, HetSANN,  SimpleHGN, and ie-HGCN. Additionally, we include the variant of ie-HGCN based on virtual nodes, ie-HGCN$_{vn}$, as a competitor.
For VN-HGCN, we employ the cross-entropy loss function, with each hidden layer, attention vector, and central node dimension set to $64$. The learning rate is set to $1e-3$, $\ell_2$ regularization parameters to $1e-4$, the number of layers to $4$, and the number of virtual nodes of each type to $16$. Dropout rate and drop-edge rate are selected via grid search from the interval $[0, 0.5]$. All parameters are optimized using Adam \cite{adam}, and VN-HGCN is trained for $1000$ epochs. The code will be made public upon acceptance of the paper.

\subsection{Experimental Results}
To compare the performance of  different methods, we conduct node classification task for all the models.
Each method is run $10$ times randomly. We perform node classification while gradually increasing the training ratio from $20\%$ to $80\%$, and report the averaged Micro-F1 and Macro-F1 in Table \ref{nc_res}. We opt not to consider runtime as an index in our experiments, aligning with the exclusion of this index in all discussed baselines except for HGT, where batch time is addressed.  

The results in Table \ref{nc_res} demonstrate that VN-HGCN achieves the best performance across most training ratios. Generally, methods that recursively aggregate neighbors outperform meta-path based methods such as metapath2vec, HAN, and MAGNN. Metapath2vec performs worse due to its inability to identify the importance between nodes and node types, while other methods employ attention mechanisms to do so. After adding virtual nodes, ie-HGCN$_{vn}$ outperforms ie-HGCN under lower training ratios, indicating that virtual nodes facilitate information flow, especially long-range message passing.

However, as the training ratio increases, the sophisticated attention structure of ie-HGCN shows better performance. Nevertheless, VN-HGCN outperforms ie-HGCN in most situations, demonstrating its flexibility in capturing the complexity of different types through long-range information. Specifically, VN-HGCN and ie-HGCN exhibit comparable performance in MicroF1 index on IMDB, with ie-HGCN marginally outperforming VN-HGCN in MacroF1 index at $60\%$ and $80\%$ training ratios. This highlights VN-HGCN's superior performance with smaller training data. Additionally, across varying training ratios, VN-HGCN outperforms all baselines in terms of both MacroF1 and MicroF1 indices for ACM and DBLP datasets ($17$ out of $24$).

\begin{table*}[htbp]
	\centering
	\caption{Experimental results for node classification task.}
	\label{nc_res}
	\resizebox{\textwidth}{!}{
		\begin{tabular}{c|c|c|c|c|c|c|c|c|c|c|c}
			\hline
			Dataset & Metrics(\%) & Training & metapath2vec & HAN & MAGNN & HGT & HetSANN & SimpleHGN & ie-HGCN & ie-HGCN$_{vn}$ & VN-HGCN \\ \hline
			& MacroF1 & 20\% & $79.20\pm.38$ & $84.32\pm.22$ & $81.18\pm.61$ & $83.70\pm.53$ & $83.94\pm.40$ & $83.51\pm.24$          & $80.53\pm.17$          & $84.44\pm.32$          & $\mathbf{86.35\pm.27}$ \\
			&         & 40\% & $81.81\pm.42$ & $84.15\pm.21$ & $82.01\pm.66$ & $84.13\pm.50$ & $85.02\pm.42$ & $84.06\pm.24$          & $85.32\pm.23$          & $85.49\pm.31$          & $\mathbf{85.51\pm.20}$ \\
			&         & 60\% & $82.14\pm.39$ & $84.71\pm.19$ & $82.33\pm.66$ & $84.87\pm.50$ & $86.63\pm.42$ & $85.46\pm.24$          & $86.51\pm.23$          & $86.57\pm.31$          & $\mathbf{86.92\pm.20}$ \\
			&         & 80\% & $82.99\pm.36$ & $85.22\pm.21$ & $83.25\pm.67$ & $85.55\pm.50$ & $86.70\pm.43$ & $85.99\pm.25$          & $85.25\pm.23$          & $\mathbf{87.81\pm.32}$ & $86.64\pm.21$ \\ \cline{2-12} 
			ACM  & MicroF1 & 20\% & $80.31\pm.37$ & $84.63\pm.22$ & $81.54\pm.63$ & $83.54\pm.53$ & $84.21\pm.41$ & $83.44\pm.25$          & $80.43\pm.18$          & $84.20\pm.33$          & $\mathbf{86.67\pm.28}$ \\
			&         & 40\% & $82.15\pm.39$ & $84.48\pm.21$ & $81.88\pm.67$ & $84.38\pm.50$ & $84.90\pm.42$ & $83.79\pm.25$          & $85.52\pm.23$          & $85.41\pm.31$          & $\mathbf{85.57\pm.20}$ \\
			&         & 60\% & $82.66\pm.38$ & $85.06\pm.19$ & $81.97\pm.69$ & $85.02\pm.50$ & $86.31\pm.42$ & $84.91\pm.24$          & $86.65\pm.23$          & $86.34\pm.31$          & $\mathbf{86.87\pm.20}$ \\
			&         & 80\% & $83.71\pm.39$ & $85.64\pm.21$ & $82.46\pm.68$ & $85.66\pm.50$ & $86.33\pm.43$ & $85.19\pm.25$          & $85.05\pm.23$          & $\mathbf{88.49\pm.32}$ & $87.06\pm.21$ \\ \hline
			& MacroF1 & 20\% & $93.53\pm.36$ & $91.69\pm.39$ & $93.13\pm.42$ & $92.14\pm.35$ & $93.13\pm.33$ & $93.17\pm.28$          & $93.62\pm.27$          & $93.57\pm.31$          & $\mathbf{93.73\pm.29}$ \\
			&         & 40\% & $92.77\pm.38$ & $91.96\pm.36$ & $93.30\pm.44$ & $92.76\pm.38$ & $93.30\pm.35$ & $93.49\pm.29$          & $94.02\pm.28$          & $93.25\pm.32$          & $\mathbf{94.09\pm.30}$ \\
			&         & 60\% & $93.59\pm.37$ & $92.14\pm.37$ & $93.84\pm.45$ & $92.47\pm.39$ & $93.84\pm.36$ & $94.12\pm.28$          & $94.69\pm.29$          & $93.52\pm.31$          & $\mathbf{94.94\pm.30}$ \\
			&         & 80\% & $93.60\pm.39$ & $92.50\pm.38$ & $93.89\pm.43$ & $94.71\pm.37$ & $93.89\pm.37$ & $95.08\pm.30$          & $95.42\pm.29$          & $94.12\pm.32$          & $\mathbf{96.09\pm.28}$ \\ \cline{2-12} 
			DBLP & MicroF1 & 20\% & $92.96\pm.35$ & $93.01\pm.40$ & $93.57\pm.41$ & $92.57\pm.36$ & $93.57\pm.34$ & $93.52\pm.30$          & $94.09\pm.29$          & $93.90\pm.32$          & $\mathbf{94.33\pm.31}$ \\
			&         & 40\% & $92.47\pm.36$ & $92.97\pm.38$ & $93.74\pm.43$ & $93.42\pm.35$ & $93.74\pm.33$ & $94.14\pm.31$          & $94.58\pm.28$          & $93.84\pm.30$          & $\mathbf{94.66\pm.30}$ \\
			&         & 60\% & $92.93\pm.35$ & $93.62\pm.39$ & $94.29\pm.42$ & $93.20\pm.34$ & $94.29\pm.37$ & $94.78\pm.31$          & $95.06\pm.28$          & $93.96\pm.30$          & $\mathbf{95.31\pm.30}$ \\
			&         & 80\% & $93.02\pm.34$ & $93.56\pm.41$ & $94.47\pm.40$ & $95.07\pm.35$ & $94.47\pm.34$ & $95.45\pm.29$          & $95.89\pm.28$          & $94.58\pm.31$          & $\mathbf{96.24\pm.30}$ \\ \hline
			& MacroF1 & 20\% & $52.55\pm.28$ & $55.47\pm.32$ & $59.35\pm.45$ & $57.95\pm.37$ & $59.35\pm.33$ & $\mathbf{59.63\pm.29}$ & $58.65\pm.30$          & $59.32\pm.32$          & $59.58\pm.29$ \\
			&         & 40\% & $54.97\pm.30$ & $62.61\pm.35$ & $60.27\pm.46$ & $60.47\pm.39$ & $61.27\pm.33$ & $64.21\pm.29$          & $63.20\pm.30$          & $64.15\pm.31$          & $\mathbf{66.36\pm.28}$ \\
			&         & 60\% & $54.28\pm.32$ & $64.58\pm.37$ & $60.66\pm.48$ & $64.85\pm.39$ & $63.66\pm.34$ & $65.73\pm.29$          & $\mathbf{67.11\pm.30}$ & $65.42\pm.31$          & $66.93\pm.30$ \\
			&         & 80\% & $56.88\pm.31$ & $63.14\pm.38$ & $61.44\pm.47$ & $67.76\pm.41$ & $64.44\pm.36$ & $66.70\pm.30$          & $\mathbf{69.53\pm.29}$ & $67.35\pm.32$          & $67.27\pm.28$ \\ \cline{2-12} 
			IMDB & MicroF1 & 20\% & $52.72\pm.27$ & $55.27\pm.30$ & $59.60\pm.46$ & $57.90\pm.36$ & $59.60\pm.35$ & $59.55\pm.30$          & $58.65\pm.31$          & $59.31\pm.33$          & $\mathbf{59.56\pm.30}$ \\
			&         & 40\% & $55.88\pm.30$ & $62.90\pm.36$ & $60.50\pm.47$ & $60.22\pm.40$ & $61.50\pm.34$ & $64.05\pm.29$          & $63.20\pm.31$          & $64.13\pm.30$          & $\mathbf{65.35\pm.28}$ \\
			&         & 60\% & $54.32\pm.32$ & $64.60\pm.37$ & $60.88\pm.46$ & $64.55\pm.38$ & $62.88\pm.35$ & $65.48\pm.30$          & $\mathbf{67.11\pm.29}$ & $65.60\pm.32$          & $66.96\pm.30$ \\
			&         & 80\% & $57.01\pm.31$ & $63.55\pm.39$ & $61.53\pm.48$ & $67.12\pm.40$ & $64.53\pm.36$ & $66.64\pm.30$          & $\mathbf{69.39\pm.29}$ & $67.42\pm.33$          & $67.76\pm.28$ \\ \hline
	\end{tabular}}
\end{table*}


\subsection{Study of the Parameters Sensitivity}
\begin{figure}
	\centering
	\includegraphics[width=\linewidth]{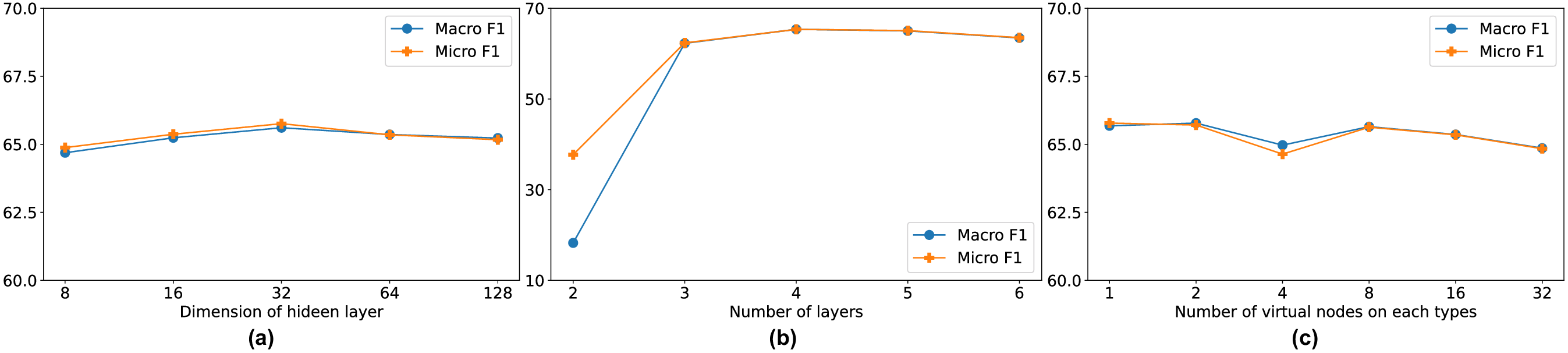}
	\caption{Classification performance of VN-HGCN w.r.t. (a). Dimension of hidden layer; (b). Number of layers; (c). Number of virtual nodes on each types ($40\%$ data for training) on IMDB dataset.}
	\label{hyper-params}
\end{figure}

In this section, we investigate the parameter sensitivity of VN-HGCN on the IMDB dataset, focusing on the dimension of the hidden layer, the number of layers, and the number of virtual nodes while keeping other parameters fixed. The results are presented in Fig. \ref{hyper-params}, from which we can derive several key observations:
\begin{itemize}
	\item {\bf Dimension of the Hidden Layer}: VN-HGCN exhibits a notable degree of robustness to alternations in the dimensionality of the hidden layer. Throughout different dimension settings, VN-HGCN consistently maintains stable performance, displaying only minor fluctuations. Nevertheless, its performance peaks when the dimensionality is configured to $32$, which strikes an effective equilibrium between having too many and too few parameters.
	
	\item {\bf Number of Layers}: The observation depicted in Fig.  \ref{hyper-params} (b) illustrates that the performance of the model improves as the network depth increases from 2 to 4 layers. However, a subsequent increase in the number of layers beyond 4 results in a drop in performance, which is consistent with the over-smoothing problem commonly encountered in overly layered models. In contrast to the standard GCN model,  VN-HGCN's constraint to $4$ layers helps alleviate the over-smoothing problem.  
	However, it's worth noting that while VN-HGCN demonstrates promise in mitigating over-smoothing, the extent to which it can completely resolve or circumvent the problem remains open. Further theoretical analysis is needed to fully understand this aspect.
	
	\item {\bf  Number of Type-level Virtual Nodes}: The sensitivity analysis demonstrates that VN-HGCN's performance remains stable across varying numbers of type-level virtual nodes. This result indicates that adding virtual nodes does not significantly impact the model's performance. Therefore, the incorporation of virtual nodes does not introduce notable changes in the model's effectiveness.
\end{itemize}
In summary, our study of parameter sensitivity reveals that VN-HGCN is robust and effective across different dimensions of the hidden layer, various numbers of layers, and type-level virtual nodes. These findings indicate the versatility and stability of VN-HGCN in handling heterogeneous graph learning tasks.

\section{Conclusion}
In this paper, we propose VN-HGCN, a novel model designed to learn node embeddings leveraging virtual nodes, enabling the aggregation of long-range information within just $4$ layers and without the need for predefined meta-paths. The key components of VN-HGCN include the allocation of type-level virtual nodes for each node type, connection via a central node to all nodes in the graph, and subsequent transformation, node-level aggregation, and calculation of inter-type attention to update node embeddings. Additionally, we demonstrate that integrating virtual nodes into other heterogeneous graph models can significantly enhance performance. Our experimental findings on node classification tasks underscore the superior performance of VN-HGCN and the effectiveness of incorporating virtual nodes. Moreover, our parameter sensitivity analysis confirms the robustness of VN-HGCN across different settings.

In conclusion, our study highlights the importance of adding virtual nodes in improving model performance, particularly in terms of how the original graph is connected. For future research directions, we plan to extend the application of our model to various tasks and explore its scalability on large-scale datasets. We believe that our work contributes novel insights into the representation of heterogeneous graph structural data and offers a simple yet powerful solution to address challenges related to over-smoothing and computational complexity. Notably, the optimal number of virtual nodes may vary across datasets, which warrants further investigation in future studies.


\printbibliography 


\end{document}